%% file: HALVersion.tex
\documentclass[fleqn,10pt]{article}
\usepackage{microtype}
\usepackage{setspace}
\usepackage{hyperref} 
\usepackage{arxiv}
\usepackage[utf8]{inputenc} 
\usepackage[T1]{fontenc}    
\usepackage{url}            
\usepackage{booktabs}       
\usepackage{amsfonts}       
\usepackage{nicefrac}       
\usepackage{microtype}      
\usepackage{lipsum}         
\usepackage{graphicx}
\usepackage{cite}
\usepackage{doi}
\usepackage{amsmath,amssymb,amsfonts}
\usepackage{algorithmic}
\usepackage{graphicx}
\usepackage{textcomp}
\usepackage{xcolor}
\def\BibTeX{{\rm B\kern-.05em{\sc i\kern-.025em b}\kern-.08emT\kern-.1667em\lower.7ex\hbox{E}\kern-.125emX}}

\usepackage{booktabs} 
\usepackage{graphicx}
\usepackage{epstopdf}
\usepackage{url}
\usepackage{algorithmic}
\usepackage{multirow}
\usepackage{caption}
\usepackage{amssymb}
\usepackage{amsfonts}
\usepackage{float}
\usepackage{bm}
\usepackage{dsfont}
\usepackage[T1]{fontenc}
\usepackage[left,modulo]{lineno}
\usepackage[ruled, vlined, linesnumbered, longend]{algorithm2e}
\usepackage{psfrag}
\usepackage{subfigure}
\usepackage{fancyhdr}
\usepackage{eucal}
\usepackage[english]{babel}
\usepackage{ifpdf}
\usepackage{cleveref} 
\usepackage{textcomp}
\usepackage{tikz}
\usepackage{tikz-inet}
\usetikzlibrary{calc,shapes.geometric}
\usetikzlibrary{shapes,snakes,arrows,automata} 
\usetikzlibrary{patterns}

\DeclareMathOperator*{\argmin}{arg\,min}
\usepackage{multirow}
\usepackage{multicol}
\usepackage{authblk}
\begin{document}
\include{macros}

\newcommand{\op}{w^+}
\newcommand{\om}{w^-}
\newcommand{\Rho}{\mathrm{P}}
\newcommand{\wresp}{\mathbf{w}^+}
\newcommand{\wresn}{\mathbf{w}^-}
\newcommand{\ve}{\mathbf{v}}
\newcommand{\noise}{\bm{\varepsilon}}
\newcommand{\idct}{\phi}
\newcommand{\chr}{\kappa}
\newcommand{\phenotype}{\chi}
\newcommand{\mapping}{\Psi}
\renewcommand{\dim}{M}
\renewcommand{\Nx}{d}
\newcommand{\State}{\mathbf{S}}
\newcommand{\pos}{\mathbf{p}}
\renewcommand{\vec}{\mathbf{v}}
\newcommand{\best}{\mathbf{b}}
\newcommand{\ti}{\!\times\!}
\newcommand{\dimwa}{\Nx\ti\Na}
\newcommand{\dimwr}{\Nx\ti\Nx}
\newcommand{\dimwfb}{\Nx\ti\Ny}
\newcommand{\dimwo}{\Nx\ti(\Na\!+\!\Ny)}
\newcommand{\espacio}[1]{}
\newcommand{\vu}{\mathbf{u}}
\newcommand{\vz}{\mathbf{z}}
\newcommand{\U}{\mathbb{U}}
\newcommand{\M}{\mathcal{M}}
\newcommand{\pred}{\hat{\mathbf{y}}}
\newcommand{\param}{\bm{\theta}}
\newcommand{\tmp}{\mathcal{T}}
\newcommand{\w}{\mathbf{w}}
\newcommand{\mlc}[1]{{\bf\color{red}#1}}
\newcommand{\msd}[1]{{\bf\color{magenta}#1}}
\newcommand{\todo}[1]{{\color{brown}\textsf{\bf [#1]}}}
\newcommand{\forceindent}{\leavevmode{\parindent=2em\indent}}
\newcommand{\fix}{\marginpar{FIX}}
\newcommand{\new}{\marginpar{NEW}}
\renewcommand{\tmp}{\mathcal{T}}
\newcommand{\goodchi}{\protect\raisebox{2pt}{$\chi$}}
\newcommand{\giv}{\,\vert\,}
\newcommand{\Space}{\mathbb{S}}
\def\rvx{{\mathbf{x}}}
\newcommand{\winunit}{{v}}
\renewcommand{\dim}{{p}}
\newcommand{\momvec}{\mathbf{m}}
\newcommand{\bo}{b^{(1)}}
\newcommand{\bt}{b^{(2)}}
\newcommand{\bj}{b^{(j)}}
\newcommand{\bJ}{b^{(J)}}
\newcommand{\uio}{u_i^{(1)}}
\newcommand{\yio}{y_i^{(1)}}
\newcommand{\zio}{z_i^{(1)}}
\newcommand{\bui}{\mathbf{u}_i}
\newcommand{\byi}{\mathbf{y}_i}
\newcommand{\bzi}{\mathbf{z}_i}
\newcommand{\buipo}{\mathbf{u}_{i+1}}
\newcommand{\byipo}{\mathbf{y}_{i+1}}
\newcommand{\wt}[1]{\widetilde{#1}}
\renewcommand\UrlFont{\color{blue}\rmfamily}

%
%
%
\renewcommand\Authfont{\bfseries}
\setlength{\affilsep}{0em}


\title{A Self-Organizing Clustering System for Unsupervised Distribution Shift Detection}
%

%
\newbox{\orcid}\sbox{\orcid}
{\includegraphics[scale=0.06]{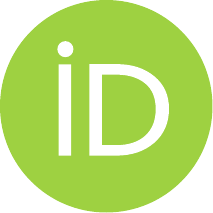}} 
\author{\href{https://orcid.org/0000-0002-9172-0155}{\usebox{\orcid}\hspace{1mm}Sebasti\'an Basterrech}}%
\author{%
\href{https://orcid.org/0000-0001-5527-5798}{\usebox{\orcid}\hspace{1mm}Line Clemmensen}
}
\affil{Department of Applied Mathematics and Computer Science, Technical University of Denmark, Denmark\\
{\{\texttt{sebbas,lkhc@dtu.dk}\}}
}
\author{
\href{https://orcid.org/0000-0002-9172-0155}{\usebox{\orcid}\hspace{1mm}Gerardo Rubino}%
}
\affil{INRIA Rennes -- Bretagne Atlantique, Rennes, France\\
{\texttt{Gerardo.Rubino@inria.fr}}
}

\date{}
\renewcommand{\headeright}{}
\renewcommand{\undertitle}{(Revised version of accepted manuscript at {IJCNN'2024})}
\renewcommand{\shorttitle}{Self-organized clustering for unsupervised shift detection}

\maketitle

\begin{abstract}
Modeling non-stationary data is a challenging problem in the field of continual learning, and data distribution shifts may result in negative consequences on the performance of a machine learning model.
Classic learning tools are often vulnerable to perturbations of the input covariates, and are sensitive to outliers and noise, and some tools are based on rigid algebraic assumptions.
Distribution shifts are frequently occurring due to changes in raw materials for production, seasonality, a different user base, or even adversarial attacks. 
Therefore, there is a need for more effective distribution shift detection techniques. 
In this article, we introduce a self-organized continual learning framework designed for monitoring and detecting distribution changes.
We explore the problem in a latent space generated by a bio-inspired self-organizing clustering and statistical aspects of the latent space.
In particular, we investigate the projections made by two topology-preserving maps: the Self-Organizing Map and the Scale Invariant Map.
Our method can be applied in both a supervised and an unsupervised context.
We construct the assessment of changes in the data distribution as a comparison of Gaussian signals, making the proposed method fast and robust.
We compare it to other unsupervised techniques, specifically Principal Component Analysis (PCA) and Kernel-PCA. Our comparison involves conducting experiments using sequences of images (based on MNIST and injected shifts with adversarial samples), chemical sensor measurements, and the environmental variable related to ozone levels.
The empirical study reveals the potential of the proposed approach.
\end{abstract}

\keywords{Distribution shift \and Continual Learning \and Topology-preserving methods \and Self Organizing Maps \and Learning Representation \and Dimensionality Reduction}

\section{Introduction}
\label{Introduction}

Robustness to data distribution shifts is a crucial design goal when developing foundation models that can be adapted to different machine learning~(ML) applications.
Classic ML approaches are still vulnerable to perturbations of the input data, and a majority of models have rigid algebraic assumptions and requires i.i.d stationary data samples~\cite{Hinder2022}. 
As a consequence, distribution shifts in the input datasets can drastically affect the model's performance.
The continual learning (CL) paradigm refers to the concept of learning a model sequentially without forgetting previously acquired knowledge, which in the context of analyzing non-stationary data has significant relevance.

In our paper, we focus on a Neural Network (NN) solution, combining the CL context with the use of non-linear dimensionality reductions based on self-organizing methods. 
We focus on Self-Organizing Maps (SOMs) for this reduction, and compare it to Scale-Invariant Maps (SIMs) (see Section~II). 
To make the proposed framework fast and efficient, we combine the SOM dimensionality reduction with a statistical analysis.
This procedure allows us to evaluate differences between probability distributions in a very efficient way.
Moreover, we don't need to assume anything about the distribution of the data.
We approximate the distribution of the latent space provided by our self-organizing clustering method through its first moments.
The efficiency in detecting changes is based on the fact that our final monitoring signal can be reasonably assumed to be Gaussian, and on the use of the 
Kullback-Leibler divergence, which is very fast to evaluate in the Gaussian case.
%

\noindent{\textbf{Contributions.}} 
We address the distribution shift problem in a context of high dimensional streaming data. 
The highlights of our contributions are the following: 
(i) We develop a bio-inspired self-organizing clustering system for assessing distribution changes in data streams. The framework can be applied in unsupervised contexts.
(ii) We investigate SOM and SIM projections, which belong to a specific family of non-linear dimensionality reduction techniques. These methods are based on a topology-preserving map, and we explore statistical aspects of the latent space. 
(iii) By construction, the proposed framework generates a univariate signal that under some restrictions (studied in the paper) can be assumed Gaussian. 

We explored the validation of our proposal over three continual learning problems.
The experimental results to support our proposal are given in Section~IV and discussed in Section~V. Section~VI presents our conclusions.

\section{Related work}
\label{RelatedWork}
\subsection{Monitoring and detection of data distribution change}

Popular approaches to monitor and assess changes in data distributions focus on the predictive accuracy of the classifier~\cite{Goldenberg2019}. In this context, the performance of various ensemble classifiers has also been studied in~\cite{Maciel:2015,Kolter:2003,du2014selective,Lapinski:2018}
The process consists in designing a classifier, and when the accuracy rate significantly decreases, then it is assumed that the data distribution has changed~\cite{Gama:2004,Goncalves2014}. However, this requires the continuous availability of ground truth labels.
Another family of techniques is based on statistical tests on raw data, such as Smirnov-Kolmogorov tests~\cite{Sobolewski:2013} and non-parametric tests~\cite{Blanco:2015,Hinder2020,Clemmensen2023data}.
Several works present an approach for monitoring aggregation metrics of the raw data, e.g. Cumulative Sum and Exponentially Weighted Moving Average. These techniques compute aggregated statistical metrics of the data and create an additive model of the first moments~\cite{Ross:2012}. For a more complete overview of the state-of-the-art in the use of data descriptors for shift detection, see~\cite{Faber21, Rabanser2019, Hinder2020, Hinder2022}.
Many distribution shift detection methods rely on the computation of the empirical estimated distributions. Even though density function estimation is a fundamental concept in statistics, the estimation of the probability mass function (pmf) is still a complex task, especially when the data belong to a high dimensional space. The approaches are sensitive to outliers and noise, and density estimation techniques may suffer from the curse of dimensionality. 
The data distribution shift can take different forms, a common taxonomy includes: sudden, gradual, and incremental shifts~\cite{Wozniak2023,SouzaChallenges2020}.
%

\subsection{Dimensionality reduction using self-organizing clustering methods}
A common approach for analyzing high-dimensional data is to apply dimensionality reduction (DR) using linear or random projections. 
However, both projections to the marginals and PCA present limitations for capturing the relevant structure of the original data that can cause data shift~\cite{Hinder2022, Ditzler2011}.
On the contrary, some authors recognize the benefits of using PCA as the DR technique in a multivariate unsupervised context~\cite{Goldenberg:2019,Qahtan2015}.
Other DR techniques, such as Kernel-PCA~\cite{Mika1998KernelPA}, scale with the number of instances. This scalability challenge makes it difficult to apply them effectively for streaming data analysis. 
An eventual approach for solving this issue is to control the number of instances in each chunk, but it will increase the hyperparameters of the model~\cite{Schoplkopf:2002}.
\noindent\textbf{Self-organizing Map (SOM)}. Also referred to as Kohonen Neural Network is a bio-inspired method, which combines concepts from Hebbian learning, vector quantization, and competitive learning~\cite{Fyfe05,Kohonen2013}.
Often, real-world data have important redundancies and intrinsic correlations between the correlated variables. 
SOMs are useful because they convert complex relationships between high-dimensional data into simple geometric relationships on a regular lattice (most often, a two-dimensional grid)~\cite{Ferles2018}.
SOMs methodology was introduced as a particular case of a Neural Network model. The method is a simple non-linear parametric mapping composed of a multi-layered network with Gaussian activation functions. The canonical case can be seen as a two-layered network, with the second layer generally  referred to as the \textit{competition layer} or \textit{feature map}.
Each neuron in the feature map is characterized by a reference weight vector, that has the same dimension as the input space.
In spite of its simplicity, the SOM algorithm is useful for DR and visualization of high-dimensional data~\cite{Saraswati2018}.
In addition, the method works well for unsupervised problems, and has the property of preserving the most important topological features of the reference data~\cite{Kohonen2013, Yin2007,Fyfe05}.
%
%
The network architecture is different from the classic feedforward case. Here, neurons are arranged on a grid. 
Each neuron~$i$ in the feature map has associated with a location in the grid that we denote to as $r_i$ and a weight vector $\w_i^{(t)}=(w_{i1}^{(t)},\ldots,w_{i\Nx}^{(t)})$. The weight vector has the dimension of the input space and connects an input pattern $\rvx^{(t)}$ with neuron~$i$.
The weights are adjusted using a physiological interpretation that considers lateral neural inhibition~\cite{Fyfe07, Kohonen01,Yin2007}. %
For simplicity, we have here omitted the reference index for the chunk.
The algorithm is iterative, composed of two main phases. 
The first phase is a competitive learning procedure, that has the goal of finding the neuron that better \textit{represents} the current input pattern.
The notion of representation is defined  by the concept of closeness among multidimensional points, where all the features of the input data have the same relevance.
The representative neuron is the unit in the feature map with associated weight vector closest to the current input sample.
At each time~$t$, in the competitive learning step, an input pattern $\rvx^{(t)}$ is presented to the network and a competition between neurons determines the \textit{representative neuron} $\winunit^{(t)}$, defined by
\begin{equation}
\label{winNeuron}
\winunit^{(t)}=\argmin_{i}{\lVert\rvx^{(t)}-\w_i^{(t)}\rVert}.
\end{equation}
In the canonical SOM, the Euclidean space was selected~\cite{Kohonen01}.
The second phase is a regression step, where the update rule for neuron~$i$ in the feature map is
\begin{equation}\label{updateRule}
\Delta\w_{i}^{(t+1)} = \eta^{(t)} h( \lVert r_{v^{(t)}} - r_i \rVert )
		\bigl( \rvx^{(t)}-\w_i^{(t)} \bigr).
\end{equation}

In this rule, $\eta(t)\in[0,1]$ is the learning rate~\cite{Smith2018} at time~$t$ and the function~$h(\cdot)$ is a smoother neighborhood function.
%
Neighborhood functions are most often chosen from the exponential family.
A typical choice is $h(z) = \exp\bigl( -z/(2\sigma^2) \bigr)$.
Often, we use $\sigma = \sigma^{(t)}$ with this function decreasing when~$t$ increases.
%
%
Its role is to weight the region of a local neighborhood centered around the representative neuron, and to control the radius of these regions.
Observe that the distance $\lVert r_{v^{(t)}} - r_i \rVert$ is defined in the grid (e.g. Manhattan).
%
%
%
Another widely selected option for a neighborhood function is the Difference of Gaussians function~\cite{Voegtlin02,MacDonald2000}.

\noindent\textbf{Scale-Invariant Map}. Some years after SOM, another self-organizing clustering named Scale Invariant Map was introduced~\cite{Fyfe05}.
SIM is also a two-layered NN, but with negative feedbacks in the activations~\cite{Fyfe05}. It was shown that this network can use a simple Hebbian learning rule for updating its weights.
Early works on SOMs applied Hebbian learning, but the resulting maps were too sensitive to the initial global parameters and showed unstable behavior~\cite{Kohonen01}.
More recent SOM specifications mitigated such behavior with the Riccati-type learning equation~\cite{Kohonen01}.
SIM seems to be a good alternative to reduce the dimensionality of the data using an essentially Hebbian learning law~\cite{Fyfe05}.
The SIM algorithm ignores the magnitude of each correlate input. 
The process responds only to the relative proportion of the magnitude of the coordinates of the input patterns~\cite{Fyfe05}. 
Another difference between SIM and the vanilla version of SOM is that before the weights are updated, in SIM the activations pass forward and backward through the neural network.
Two possible criteria exist for selecting the representative neuron. One criterion selects the neuron with the greatest activation. 
Another criterion is given in Expression~(\ref{winNeuron}).
In this work, we only investigate the latter. 
The results presented in~\cite{Fyfe05} show how a Hebbian learning rule leads to the following weight update for node~$i$:
\begin{equation}
\label{SIMupdate}
\Delta\w_{i}^{(t+1)} = \eta^{(t)} h( \lVert r_{v^{(t)}} - r_i \rVert )
		\bigl( \rvx^{(t)}-\w_{v^{(t)}}^{(t)} \bigr),
\end{equation}
where $v^{(t)}$ is a representative neuron in the feature map, and $h(\cdot)$ is the transfer function (often the same Difference of Gaussians function as in SOMs). 
Note that, according to Expression~(\ref{SIMupdate}), the representative neuron has a direct effect over the other weights. 

This is a difference with SOM, and impacts the tool in how the map creates the clusters. A trained SOM approximates a Voronoi tessellation of the input space, while SIM makes a mapping forming a kind of ``pie chart'' where each neuron represents a slice of the input data~\cite{MacDonald1999}. 

\section{Methodology}
\subsection{Problem formulation} CL in real-world applications is usually associated with time-dependent problems in a non-stationary environment~\cite{ConceptAdaptationSurvey}. Therefore, we incorporate time indexed over a discrete set~$\tmp$ (typically, a segment of integers) into our notation.
In this context, we consider a data stream as an ordered sequence of chunks $\{S_1, S_2,\ldots, S_i, \ldots\}$, where chunk~$S_i$ consists of a finite data sequence $S_i=\{\rvx_i^{(1)},\rvx_i^{(2)}, \ldots, \rvx_i^{(K)}, \ldots\}$ containing realizations of a probability measure, valued in some data space~$\goodchi$.
We assume that we have a Markov kernel process from $\tmp$ to {$\goodchi$} that associates a distribution $p_t$ of $\goodchi$ with every time point $t\in \tmp$.
A data shift occurs when there exist at least two time stamps $t_i$ and $t_j$ such that $p_{t_i}$ and $p_{t_j}$ are different enough~\cite{Webb:2018}.
The goal in a data distribution shift detection problem is to identify all points in time~$t$ in~$\tmp$ such that~$p_t$ and~$p_{t+\Delta}$ differ significantly over a small-time interval~$\Delta>0$.
%
%
The magnitude of such difference can be estimated using a dissimilarity metric between the underlying distributions~\cite{Hinder2022}. 
Once the shift is quantified, then a decision rule is applied for deciding if the dissimilitude is either significant or not.

\subsection{Unsupervised dimensionality reduction using topographic maps} 
Comparing probability distributions is a fundamental and difficult problem in statistics, and it is particularly challenging in large dimensions.
Therefore, several shift detection techniques in the unsupervised context rely on projecting the data into a latent space. 
Typical choices are projections over the coordinate axes (marginals), random projections and projections onto the principal components~\cite{Hinder2022}.
Linear descriptors of the data present well-known limitations~\cite{Qahtan2015}.
Other nonlinear projections, such as 
Kernel-PCA~\cite{Schoplkopf:2002}, autoencoders, and t-SNE~\cite{vanDerMaaten2008}, reduce the dimensionality of the data while preserving the reference structure.
However, they may present some disadvantages for being used for streaming data analysis. Kernel-PCA scales with the number of samples. Large NNs and autoencoders require additional time windows, and the number of required samples grows with the number of dimensions.

\begin{figure}[htbp]
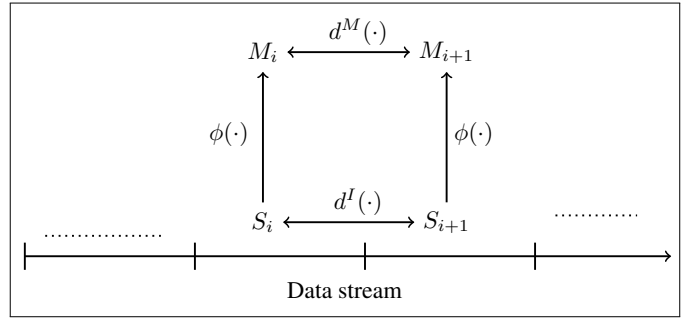

\centering
\scalebox{0.9}{
\fbox{
\tikz{
\node[rotate around={0:(0,0)},thick, minimum size=3.5ex]  (a0) at (4.9,3.3) {$d^{M}(\cdot)$};
\node[rotate around={0:(0,0)},thick, minimum size=3.5ex]  (a0) at (4.9,0.8) {$d^I(\cdot)$};
\node[rotate around={0:(0,0)},thick, minimum size=3.5ex]  (m0) at (3.5,3) {$M_i$};
\node[rotate around={0:(0,0)},thick, minimum size=3.5ex]  (m1) at (6.2,3) {$M_{i+1}$};
\node[rotate around={0:(0,0)},thick, minimum size=3.5ex]  (f0) at (3,1.8) {$\phi(\cdot)$};
\node[rotate around={0:(0,0)},thick, minimum size=3.5ex]  (f0) at (6.6,1.8) {$\phi(\cdot)$};

\node[rotate around={0:(0,0)},thick, minimum size=3.5ex]  (id0) at (3.5,0.5) {$S_i$};
\node[rotate around={0:(0,0)},thick, minimum size=3.5ex]  (id1) at (6.2,0.5) {$S_{i+1}$};
%
\node[thick, minimum size=3.5ex]  (id2) at (4.7,-0.5) {Data stream};
\draw[-,black] (7.8,0.6)  edge[dotted,thick] (9,0.6);
\draw[-,black] (id0)  edge[->,thick] (m0);
\draw[-,black] (m1)  edge[<->,thick] (m0);
\draw[-,black] (id0)  edge[<->,thick] (id1);
\draw[-,black] (id1)  edge[->,thick] (m1);
\draw[-,black] (0.3,0.3)  edge[dotted,thick] (2,0.3);
\draw[-,black] (0,-0.2)  edge[-,thick] (0,0.2);
\draw[-,black] (5,-0.2)  edge[-,thick] (5,0.2);
\draw[-,black] (7.5,-0.2)  edge[-,thick] (7.5,0.2);
\draw[-,black] (2.5,-0.2)  edge[-,thick] (2.5,0.2);
\draw[-,black] (0,0) edge[->,thick] (9.5,0);
%
}}}
\caption{Visualization of the building descriptor of the data. The distance $d^{I}(\cdot)$ computes a similarity between two distributions directly from the raw data. On the other hand,
the distance $d^{M}(\cdot)$ computes a similarity between two distributions in the latent space. We denote the non-linear projections using a topographic map $\phi(\cdot)$.}
\label{Descriptor}  
\end{figure}

Here, we investigate another approach that consists of using a trained self-organizing clustering method to make the transformation from the input space to the latent space. In the empirical analysis, we study both SOM and SIM techniques.
The training phase is made using an initial time-window of the data stream.
Then, we continue the learning as usual in a CL scenario.
Figure~\ref{Descriptor} presents a basic approach for tracking changes in the distribution shift. 
This strategy is based on the assumption that performing a similarity analysis directly in the input space may be too expensive. 
Consequently, it is more computationally efficient to transform the data in a latent space and, only then, to make the similarity analysis.
%

%
\subsection{Embedding procedure for the distance matrix} 
Both SOMs and SIMs have frequently been applied as an unsupervised  tool for clustering problems~\cite{Kohonen01}. Also, they showed to be effective for DR.
Both methods project the input into the feature map, which it is most often a 2-dimensional space, composed of the coordinates of the neurons. 
Therefore, it is a powerful reduction that can be useful in many applications, but for detecting distribution changes in the input space such a dramatic reduction of the original information may have negative consequences.
In addition, projection over coordinates has other drawbacks. Coordinates are arbitrarily selected, and more often they don't consider any property of the data itself. 
There are even problems in cases where the coordinates are not natural in any sense~\cite{Carlsson2006}.

\begin{figure}
    \centering
\includegraphics[scale=0.2]{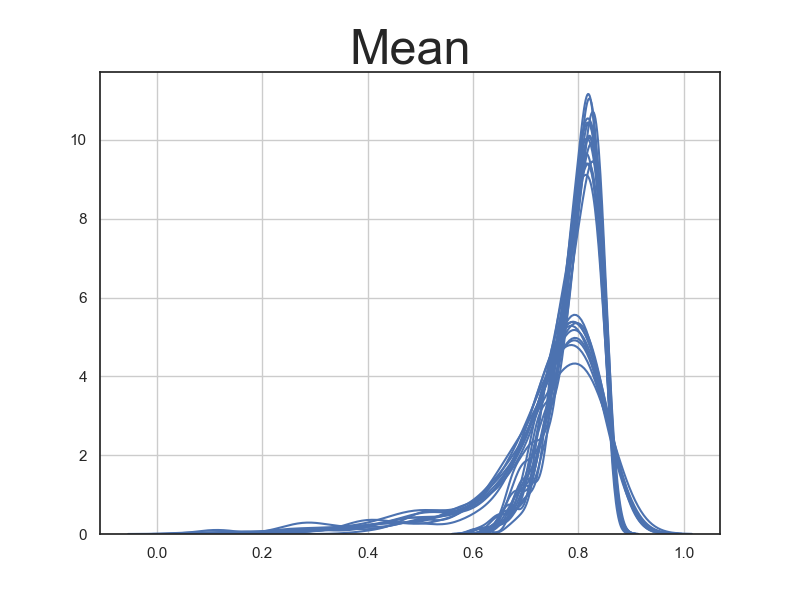}
\includegraphics[scale=0.2]{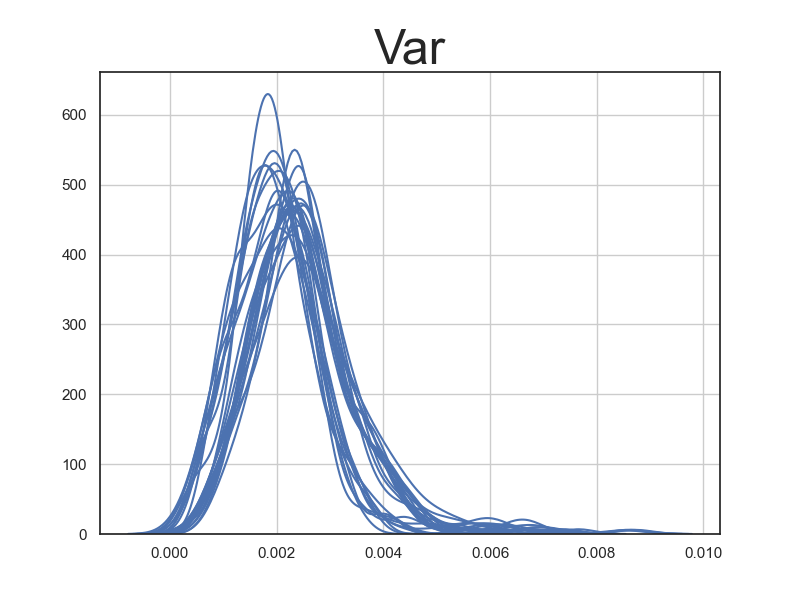}
\includegraphics[scale=0.2]{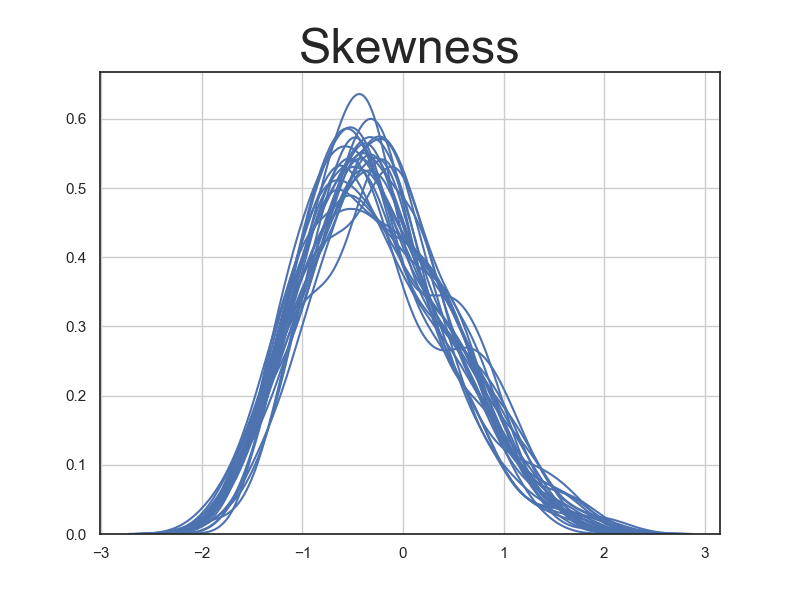}
\includegraphics[scale=0.2]{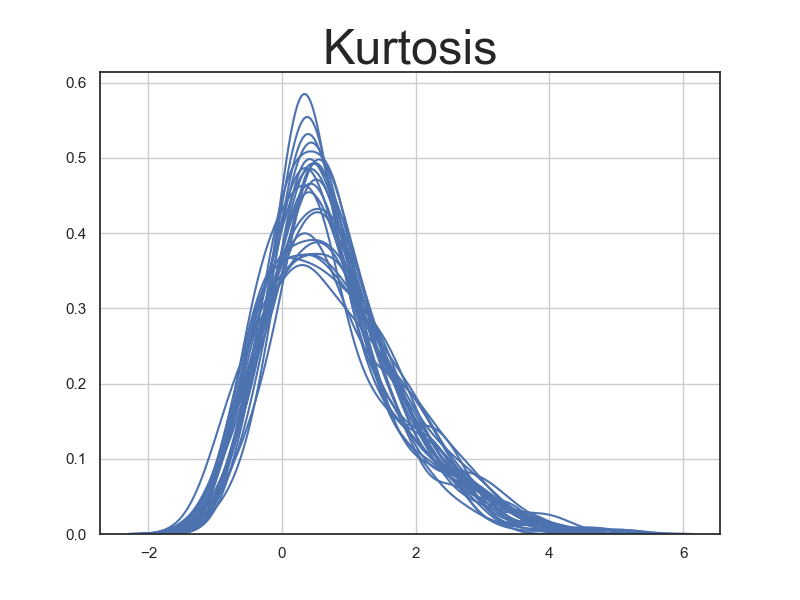}
    \caption{Visual comparison of the significance of each of the moments for representing the information in the matrix $D$. The data corresponds to the problem of MNIST with adversarial samples in a CL setting.}
    \label{Moments}
\end{figure}

To overcome these difficulties, we decided to use a less rigid projection that is coordinate-free and also that contains more information from the original space.
Instead of projecting over two dimensions, we propose a reduction to a latent space in $\dim^2$ dimensions, where $\dim$ is the number of neurons in the feature map. 
Hence, our focus is on the geometric properties of the latent space. We use the way the lattice grid is defined, where each neuron in the grid is characterized by its weight vector and the location in the grid.
%
%
Once the model is trained, we construct a matrix~$D$, where the element in the position $(i,j)$ is the distance between the input data and the weights of the neuron located at the position $(i,j)$. 
In other words, $d_{i,j}=\lVert \rvx - \w_{k}\rVert_2$, where $\w_k$ is the weight vector associated with the $(i,j)$-neuron.
%
%
We encode the information in matrix $D$ with dimensions~$\dim^2$, that is still  large. Note that, standard empirical experiments use a squared grid of few hundred of neurons~\cite{Kohonen2013}.
With the nonlinear projection of the topographic map, the dimension is reduced from $\Nx$ to $\dim^2$, which may still be too large for estimating densities. Then, we apply an additional dimensionality reduction step by moving to the first moments of the distribution represented in~$D$.
%
%
The first moments of a random sequence capture different aspects of the probability distribution that are useful to approximate it~\cite{Chamseddine2012}.
%
%
To evaluate the relevance of the moments in our problem, we empirically analyzed the impact of the first four moments.
Let us denote $\momvec=(m_1,m_2,m_3,m_4)\in\R^{4}$ these moments, where $m_1$ is the mean, $m_2$ the variance, $m_3$ the skewness, and $m_4$ the kurtosis.
%
%
%
Figure~\ref{Moments} presents an example of  the information provided by each of the first four moments. 
The analysis was made offline, comparing the samples from the MNIST benchmark data versus the adversarial MNIST samples (for a description of the benchmark data see Section~\ref{BenchmarkData}). 
Figure~\ref{Moments} has four plots, each having information about the mean, variance, skewness and kurtosis. 
Each curve represents the estimated pmf of the sequence $\momvec_i^{(1)},\ldots,\momvec_i^{(K)}$, for all chunks $S_i$.
According to this chunk-based analysis, it is visible that the left plot in Figure~\ref{Moments} distinctly reveals two families of densities.  
One group of curves represents the pmf calculated using the original MNIST samples, while the other group corresponds to the pmf computed with the adversarial MNIST samples.
This shows that the mean captures \textit{sufficient} information from matrix~$D$ to be able to distinguish groups of data distributions. Therefore, the proposed framework includes a final dimensionality reduction step, which involves transforming the entire matrix~$D$ into a univariate sequence. Then, we represent~$D$ using the first moment (mean function) of its~$\dim^2$ elements (seen as $\dim^2$ independent samples of an auxiliary random variable).

\subsection{Workflow of the proposed framework}
\begin{figure}[htbp]
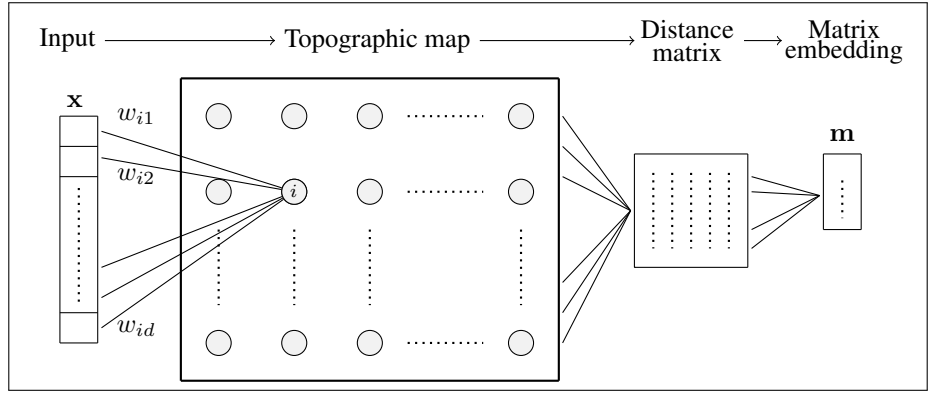

\centering
\fbox{
\scalebox{1}{
\tikz{
\node (t0) at (-1,2) {Input};
\node (t1) at (3.1,2) {Topographic map};
\node (t2) at (7.2,2.15) {Distance};
\node (t3) at (7.2,1.85) {matrix};
\node (t4) at (9.275,2.15) {Matrix};
\node (t5) at (9.275,1.85) {embedding};
\draw (t0)  edge[->,black] (t1);
\draw (t1)  edge[->,black] (6.5,2);
\draw (7.95,2)  edge[->,black] (8.45,2);
\node[circle,draw, fill=gray!10,minimum size=4.5] (n0) at (1,0) {};
\node[circle,draw, fill=gray!10,minimum size=4.5] (n1) at (2,0) {};
\node[circle,draw, fill=gray!10,minimum size=4.5] (n2) at (3,0) {};
\node[circle,draw, fill=gray!10,minimum size=4.5] (n3) at (5,0) {};
\node[circle,draw, fill=gray!10,minimum size=4.5] (n4) at (1,-2) {};
\node[circle,draw, fill=gray!10,minimum size=4.5] (n5) at (2,-2) {};
\node[circle,draw, fill=gray!10,minimum size=4.5] (n6) at (3,-2) {};
\node[circle,draw, fill=gray!10,minimum size=4.5] (n7) at (5,-2) {};
\node[circle,draw, fill=gray!10,minimum size=4.5] (n8) at (1,1) {};
\node[circle,draw, fill=gray!10,minimum size=4.5] (n9) at (2,1) {};
\node[circle,draw, fill=gray!10,minimum size=4.5] (n10) at (3,1) {};
\node[circle,draw, fill=gray!10,minimum size=4.5] (n11) at (5,1) {};
\node[circle,draw, fill=gray!10,minimum size=4.5] (n12) at (2,0) {};
\node (n1) at (2,0.01)  {$_i$};
\draw[-,black] (3.5,0)  edge[dotted,thick] (4.5,0);
\draw[-,black] (3.5,1)  edge[dotted,thick] (4.5,1);
\draw[-,black] (3,-0.5)  edge[dotted,thick] (3,-1.5);
\draw[-,black] (1,-0.5)  edge[dotted,thick] (1,-1.5);
\draw[-,black] (2,-0.5)  edge[dotted,thick] (2,-1.5);
\draw[-,black] (5,-0.5)  edge[dotted,thick] (5,-1.5);
\draw[-,black] (3.5,-2)  edge[dotted,thick] (4.5,-2);
\draw[-,black] (0.5,1.5)  edge[thick] (0.5,-2.5);
\draw[-,black] (0.5,1.5)  edge[thick] (5.5,1.5);
\draw[-,black] (0.5,-2.5)  edge[thick] (5.5,-2.5);
\draw[-,black] (5.5,1.5)  edge[thick] (5.5,-2.5);
\draw[-,black] (-1.1,1)  edge (-0.6,1);
\draw[-,black] (-1.1,-2)  edge (-0.6,-2);
\draw[-,black] (-1.1,1)  edge (-1.1,-2);
\draw[-,black] (-0.6,1)  edge (-0.6,-2);
\node (i0) at (-0.9,1.2) {$\mathbf{x}$};
\draw[-,black] (-1.1,0.6)  edge (-0.6,0.6);
\draw[-,black] (-1.1,0.2)  edge (-0.6,0.2);
\draw[-,black] (-1.1,-1.6)  edge (-0.6,-1.6);
\draw[-,black] (-0.85,0.05)  edge[dotted,thick] (-0.85,-1.5);
\draw[-,black] (-0.55,0.45)  edge (n12);
\draw[-,black] (-0.55,0.8)  edge (n12);
\draw[-,black] (-0.55,-1.)  edge (n12);
\draw[-,black] (-0.55,-1.4)  edge (n12);
\draw[-,black] (-0.55,-1.8)  edge (n12);
\node (i1) at (-0.09,1) {$w_{i1}$};
\node (i6) at (-0.09,0.2) {$w_{i2}$};
\node (i6) at (-0.09,-1.8) {$w_{id}$};
\draw[-,black] (6.5,-1)  edge (6.5,-1);
\draw[-,black] (6.5,0.5)  edge (6.5,-1);
\draw[-,black] (8,0.5)  edge (8,-1);
\draw[-,black] (6.5,0.5)  edge (8,0.5);
\draw[-,black] (6.5,-1)  edge (8,-1);
\draw[-,black] (6.75,0.25)  edge[dotted,thick] (6.75,-0.75);
\draw[-,black] (7,0.25)  edge[dotted,thick] (7,-0.75);
\draw[-,black] (7.25,0.25)  edge[dotted,thick] (7.25,-0.75);
\draw[-,black] (7.5,0.25)  edge[dotted,thick] (7.5,-0.75);
\draw[-,black] (7.75,0.25)  edge[dotted,thick] (7.75,-0.75);
\draw[-,black] (5.55,1)  edge[-] (6.45,-0.25);
\draw[-,black] (5.55,0.6)  edge[-] (6.45,-0.25);
\draw[-,black] (5.55,0.2)  edge[-] (6.45,-0.25);
\draw[-,black] (5.55,-2)  edge[-] (6.45,-0.25);
\draw[-,black] (5.55,-1.6)  edge[-] (6.45,-0.25);
\draw[-,black] (5.55,-1.2)  edge[-] (6.45,-0.25);
\draw[-,black] (9,0.5)  edge (9,-0.5);
\draw[-,black] (9.5,0.5) edge (9.5,-0.5);
\draw[-,black] (9,0.5) edge (9.5,0.5);
\draw[-,black] (9,-0.5) edge (9.5,-0.5);
\draw[-,black] (8.05,0)  edge[-] (8.95,-0.05);
\draw[-,black] (8.05,0.2)  edge[-] (8.95,-0.05);
\draw[-,black] (8.05,-0.75)  edge[-] (8.95,-0.05);
\draw[-,black] (8.05,-0.5)  edge[-] (8.95,-0.05);
%
%
\draw[-,black] (9.25,0.15)  edge[dotted,thick] (9.25,-0.35);
\node (m) at (9.25,0.75) {$\momvec$};
}}
}
\caption{Visualization of the proposed approach.}
\label{ProposedApproach}  
\end{figure}

Figure~\ref{ProposedApproach} illustrates the main three-stages of the proposed framework for monitoring distribution variation in the data stream:
\begin{enumerate}
    \item[(i)] \textbf{Dimensionality reduction}. A non-linear projection of the high-dimensional input pattern is applied. The projection is made using a topology-preserving mapping. 
    \item[(ii)] \textbf{Clustering analysis}. A matrix is created that displays the geometric relations between the projected points and their distances to the representative neurons of each cluster. Note that, this study is made in the latent space that has much lower dimensionality than the original space.
    \item[(iii)] \textbf{Statistical summary}. The distance matrix is embedded using a statistical summary. In this work, we present experiments with the matrix $D$ summarized with the mean function. Then, the framework provides a univariate signal for monitoring the changes in the distribution.
    \item[(iv)] \textbf{CL step - Model update}. Each representative neuron is characterized by its weight vector, that is trained using an initial window of samples. In case of detecting a significant change in the distribution, the  weights are updated with the data presented in the last seen chunk.
\end{enumerate}

\renewcommand{\Y}{\mathcal{Y}}
\renewcommand{\vu}{\momvec}
\subsection{Quantification of the distribution shift}
\label{Quantification}
%
Observe first that the distribution of the mean of a sequence, as the number of terms increases, tends toward a Gaussian under conditions often satisfied in practice (Central Limit Theorem, CLT). 
We need enough terms (a few dozens is in general enough) and independence between them, or at least a weak correlation. Assuming this holds will simplify the computations that we describe in the following lines.
The number of terms for satisfying the CLT is given by the number of neurons in the SOM/SIM lattice.
We compare the data distributions in two consecutive chunks $S_i$ and $S_{i+1}$, using the Kullback-Leibler (KL) divergence. 
The KL-divergence method has been employed to monitor alterations in the data distribution~\cite{Basterrech2022SMC}. 
As illustrated in Figure~\ref{Descriptor}, the KL divergence score is calculated in the projected space, rather than making the assessment directly on the original data.
Given two pmfs~$p$ and~$q$, defined in a common data space~$\goodchi$, the KL-divergence from~$p$ to~$q$ is~\cite{Dasu2006}
\begin{equation}
\label{KLformule}
    {\rm{KL}}(p\parallel q)=\ds{\sum_{s\in\Space} p(s) \log\frac{p(s)}{q(s)}}.
\end{equation}
This is not strictly a distance. It does not satisfy the triangular inequality and is not symmetric in inputs.
The $\rm{KL}(p \parallel q)$ quantifies the information lost if we use~$q$ as an approximation of~$p$.
%
%
Even though the KL-divergence is not strictly a distance, it has several useful properties and advantages over mathematical distances (for more details, see~\cite{Cover2012, Dasu2006,Basterrech2023ISDA}).
One of the benefits of KL is that there exists a relationship with the expected value of the likelihood ratio. Moreover, Expression~(\ref{KLformule}) takes a specific form for specific distributions. 
This is the case when two Gaussian distributions are compared. Let $\mu_p$ and $\sigma^2_p$ (resp. $\mu_q$ and $\sigma^2_q$) be the mean and variance of the pmf $p$ (resp. $q$).
In this case, we have
\begin{equation}
\label{KLformuleGauss}
    {\rm{KL}}(p\parallel q)=
    \ds{
    \log\frac{\sigma_q}{\sigma_p}+ 
    \frac{\sigma_p^2+(\mu_p-\mu_q)^2}{2\sigma_q^2}-\frac{1}{2}
    }.
\end{equation}

\subsection{Decision rule module}
%
The framework has an additional independent block with the decision rule.
Here, we work with a simple and fast procedure taken from the area of time series analysis for detecting outliers.
Let $\{l_1,l_2,\ldots \}$ be the signal with the KL-divergence scores. This signal is computed as described in Section~\ref{Quantification}.
We apply a rule that is often used in EEG analysis to detect artifacts~\cite{BasterrechNCAA2019}. We define a \textit{critical} point location (i.e., timestamps when the distribution change is relevant) when 
$l_i\notin[\bar{l}-\alpha\sigma,\bar{l}+\alpha\sigma]$,  where $\bar{l}$ denotes the mean of the sequence, $\sigma$ is the standard deviation, and $\alpha$ is a real-value control parameter.

%

\section{Results}
\subsection{Datasets}
\label{BenchmarkData}

There is a lack in the community of a large and diverse collection of real data streams in a high-dimensional space~\cite{SouzaChallenges2020}, especially in unsupervised streaming data analysis.
As a consequence, we generated a synthetic data stream with injected drifts. The created stream has samples from MNIST~\cite{lecun2010} and adversarial samples of MNIST data generated in~\cite{Rabanser2019}.
Furthermore, we carried out experiments over other two real-world datasets.
%

\begin{figure}
    \centering
    \includegraphics[scale=0.14]{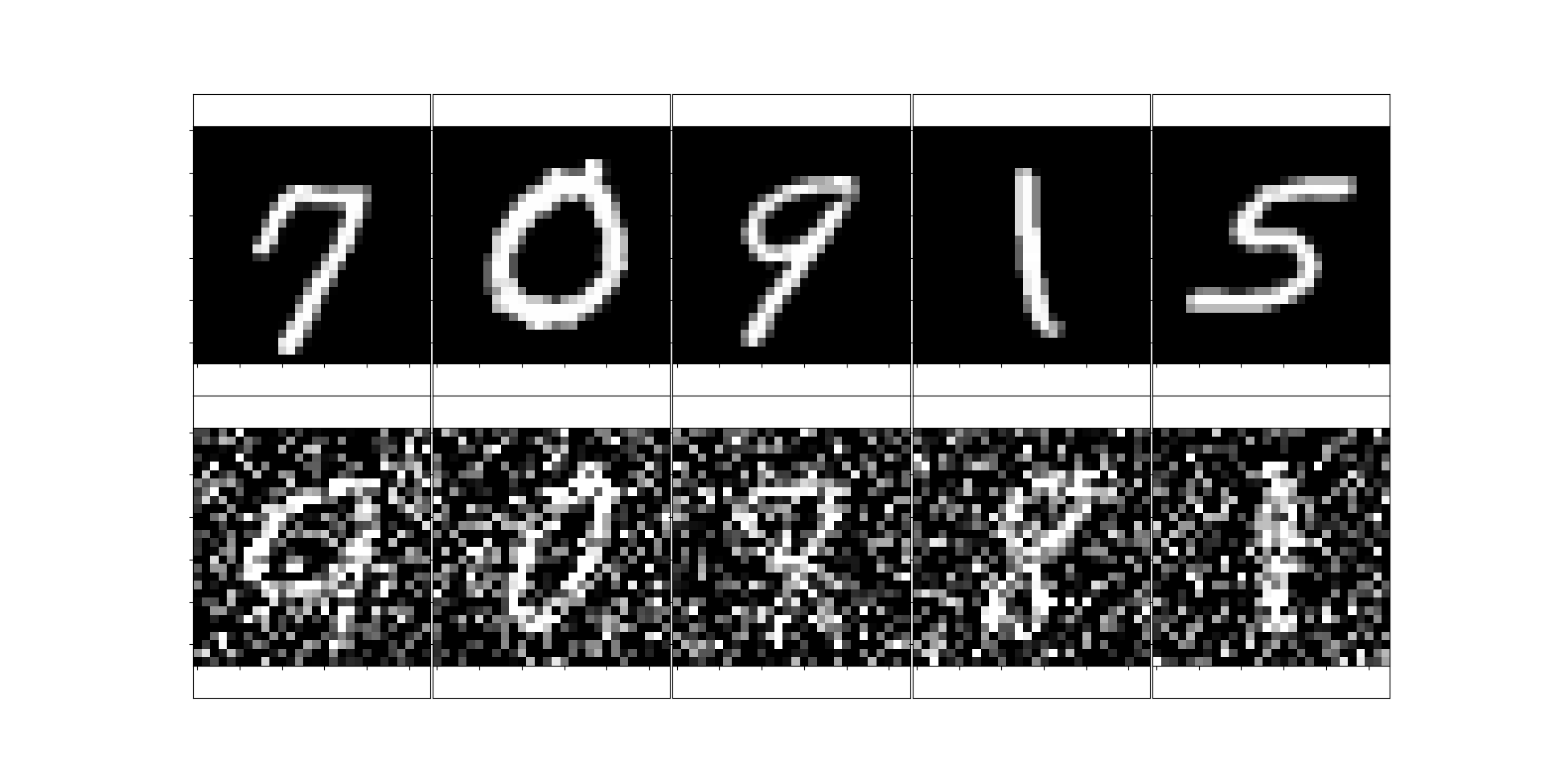}
\centering
\scalebox{0.7}{
\tikz{
\node[rotate around={50:(0,0)},thick, minimum size=3.5ex]  (id1) at (1.2,1.35) {MNIST dataset};
\node[rotate around={50:(0,0)},thick, minimum size=3.5ex]  (id1) at (3.5,1.3) {adversarial MNIST};
\node[rotate around={50:(0,0)},thick, minimum size=3.5ex]  (id1) at (6.2,1.35) {MNIST dataset};
%
\node[thick, minimum size=3.5ex]  (id1) at (4.2,-0.5) {Data stream};
\draw[-,black] (7.8,0.6)  edge[dotted,thick] (9,0.6);
\draw[-,black] (0,-0.2)  edge[-,thick] (0,1);
\draw[-,black] (5,-0.2)  edge[-,thick] (5,1);
\draw[-,black] (7.5,-0.2)  edge[-,thick] (7.5,1);
\draw[-,black] (2.5,-0.2)  edge[-,thick] (2.5,1);
\draw[-,black] (0,0) edge[->,thick] (9.5,0);
%
}}
\caption{\label{CIFER10-Example}MNIST problem with adversarial samples: This example illustrates the transition between the sequence of images before and after the injected drift. The second row of images contains the sequence of adversarial samples.}
\end{figure}

\noindent\textbf{MNIST with adversarial samples}~\cite{Rabanser2019}. We created a data stream, emulating recurrent data shifts. It contains original MNIST images and adversarial images created via FGSM. For more details about the adversarial MNIST data see~\cite{Rabanser2019}.
Figure~\ref{CIFER10-Example} depicts how the streaming data was generated. The construction of the data stream simulates changes in distribution when the datasets are exchanged from original samples to fake samples. A similar method of data stream generation was used in the context of fake information classification~\cite{Basterrech2023DSAA}. It has inter-exchanged samples from both datasets. Every 2000 images, we exchange the datasets. We assume that an efficient data shift monitoring will detect the time-stamps where there are changes between regular MNIST samples and adversarial MNIST samples.
~\\
\noindent\textbf{Gas sensor data stream}~\cite{Vergara2012}. 
This dataset was initially proposed as a classification benchmark problem with 16 sensors monitoring metal-oxide gas over three years, and where sensor drifts occur and deteriorates classifications.
%
%
During the last years, the dataset has become popular in the domain of concept drift detection. 
Here, we analyze an updated version with~128 features and six classes representing gaseous substances~\cite{Vergara2012,SouzaChallenges2020}. 
The number of changes is~38.3 shifts according to~\cite{Souto:2008}.
~\\
\noindent\textbf{Ozone data stream}~\cite{SouzaChallenges2020}.
This dataset contains air measurements collected from 1998 to 2004. It has~72 features and a binary output variable  (ozone day and normal day). The problem has only~2534 samples, for more details see~\cite{Effrosynidis2021,SouzaChallenges2020}.
It is considered a challenging problem because the data is imbalanced. In addition, it \textit{seems} that they have a high frequency of distribution changes~\cite{SouzaChallenges2020}.
%
%

\subsection{Experimental setup}
The three studied benchmark problems can be analyzed in a supervised context. 
However, here we make the monitoring of the data distributions only by analyzing the covariate variables.
%
For each of the problems, we used the first $30\%$ of samples for training the parameters of the self-organizing clustering methods (SOM and SIM).
This training was made offline, as a pre-phase of the continual learning process. 
Then, we simulated a streaming environment using the \texttt{scikit-multiflow} package~\cite{skmultiflow}. 
Both SOM and SIM have a grid with $10\times 10$ neurons.
The two parameters of the decision rule, $\alpha$ and window size, were analyzed on a grid, $\alpha$ in $[1,29]$ and window size in $[1,25]$, only considered the even values.

We generated 30 data streams using the MNIST and adv. MNIST datasets. The chunk size parameter was studied in the set $\{50,100,200,500\}$.
For the MNIST problem, in case of a chunk size equal to 200, the stream has simulated sudden drifts. The other chunk values create chunks with original MNIST data and fake samples (both in the same chunk).

\noindent\textbf{Metric choices and baseline}. As a way to assess the quality of detecting distribution shifts, we compute the Kappa score for the generated MNIST data stream. The Kappa score is commonly used to compare binary sequences.
The data stream was synthetically created, so we have the precise time-stamps wherein the shifts occur.
We also use the same MNIST data to evaluate PCA and Kernel-PCA as drift detector tools.
We use those two classic clustering techniques as a baseline.
We assess the capacity of PCA and Kernel-PCA as follows. 
We utilize an initial time window comprising~$30\%$ of the data stream for training both clustering methods.
Subsequently, we apply the trained methods to project the data points into a latent space, following a CL setting.
Once the data in the current chunk are projected, a pmf is estimated using the projected points.
Subsequently, comparisons between pmfs were conducted in the latent space using two popular metrics~\cite{Hinder2022}: the cumulative histogram and the Kolmogorov-Smirnov test.
The monitoring of the magnitude of the shifts was done with KL-divergence.

\subsection{Empirical evaluation}

\begin{figure}
    \centering
\includegraphics[scale=0.5]{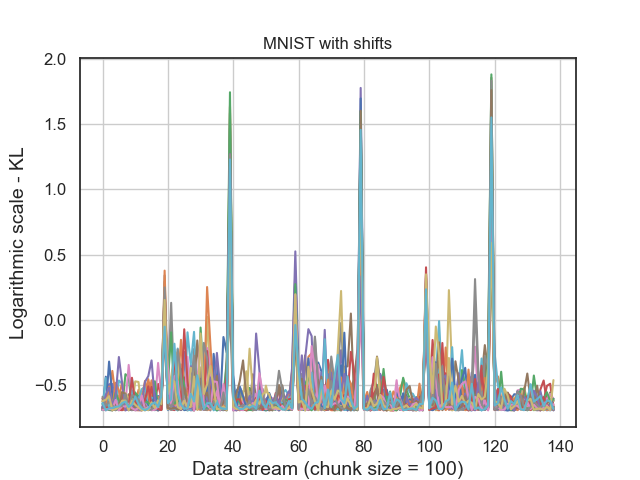}
\includegraphics[scale=0.5]{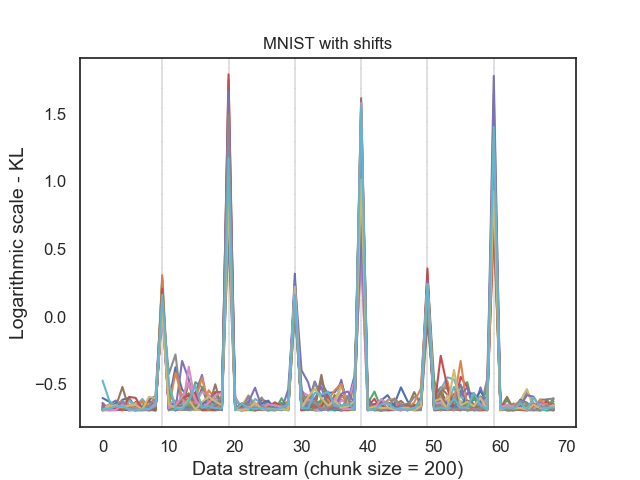}
    \caption{Analysis of distribution shifts with fixed reference time windows: Each curve represents the shift monitoring for a specific generated data stream. We generated 30 data streams using the MNIST dataset and injected shifts using adversarial samples. The left figure was generated with data streams using chunk size 100 samples. The right figure was generated with data streams using chunks with 200 samples.}
    \label{exampleMNIST}
\end{figure}

\begin{figure}
    \centering
\includegraphics[scale=0.4]{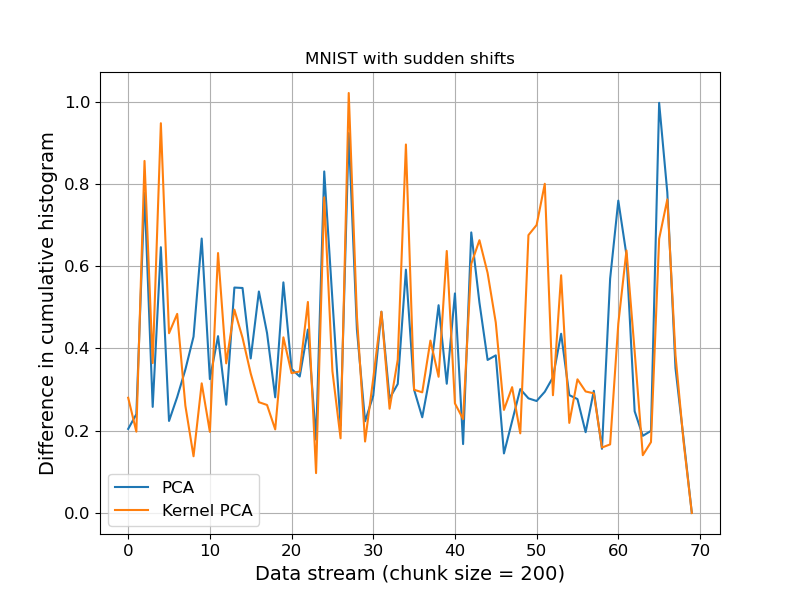}
\includegraphics[scale=0.4]{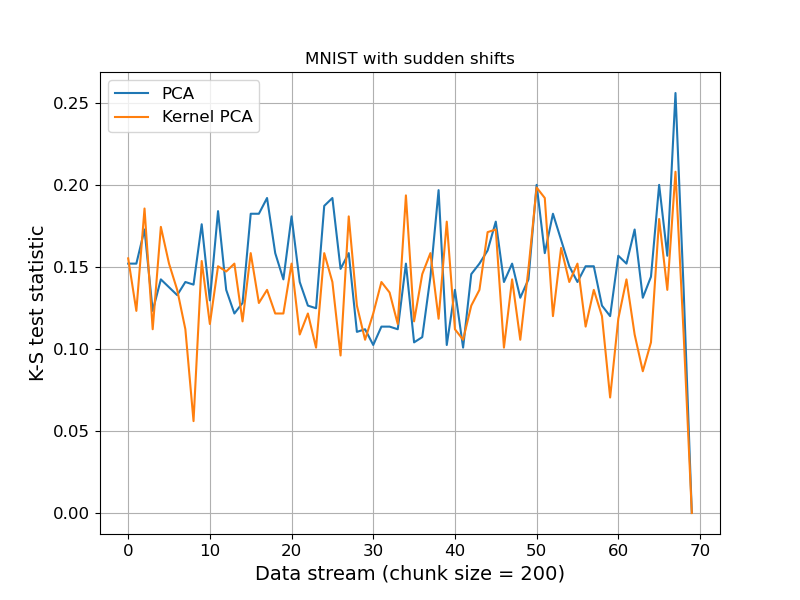}
\caption{Example of the difficulty of the problem. The projection was done with PCA and Kernel-PCA. The left figure shows the difference between two cumulative histograms computed over two consecutive chunks.The right figure shows the statistic of Kolmogorov-Smirnov test over two consecutive chunks.}
    \label{KPCA-PCA}
\end{figure}

Figure~\ref{exampleMNIST} presents the monitoring of the MNST data stream. There are~$30$ curves per graph, with each curve representing the results from different data streams.
The left curve displays the results when the number of samples in the chunk was 100 (with injected shifts occurring at time steps~$20$, $40$, $60$, $80$, $100$, and $120$). Conversely, the right curve illustrates the results when the number of samples was~$200$, with injected shifts in time-stamps~$\{10,20,30,40,50,60\}$.
Both figures show how the proposed framework effectively monitors changes in the distribution.
Contrarily, Figure~\ref{KPCA-PCA} illustrates the similarity between two consecutive chunk distributions when the baseline methods (PCA and Kernel-PCA) were applied (with a chunk size equal to 200).
The left figure presents the results of assessing the difference between the cumulative histograms in the latent space (with four principal components). 
The right figure presents the results of applying the Kolmogorov-Smirnov test to measure the discrepancy between the cumulative histograms.
Both graphics show the complexity of the problem, and how PCA and Kernel-PCA have limitations for detecting fine types of drifts in the data.

\begin{figure}
    \centering
{\includegraphics[scale=0.6]{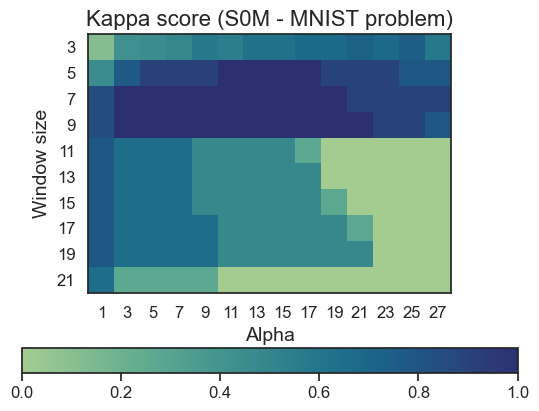}}
{\includegraphics[scale=0.6]{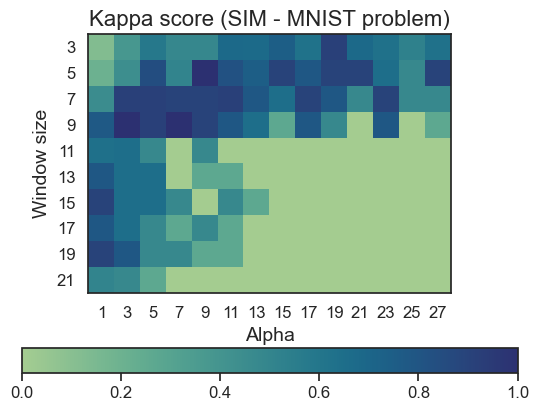}}
    \caption{Analysis of sensitivity regarding the two parameters used in the decision rule (the $\alpha$ rate and the window size) was conducted. The accuracy of shift prediction was assessed using the Kappa score.}
    \label{Kappa}
\end{figure}

\begin{figure}
    \centering
{\includegraphics[scale=0.6]{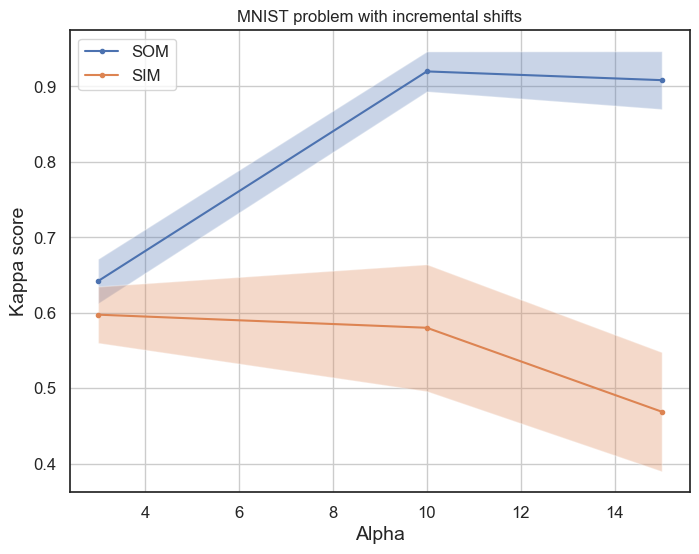}}
    \caption{Incremental shift. Analysis of sensitivity of the two parameters used in the decision rule. The data stream has the MNIST with adversarial samples, and the shift is injected incrementally. The CI was computed with the results of 30 different generated data streams.}
    \label{Incremental}
\end{figure}

\begin{figure}
    \centering
{\includegraphics[scale=0.7]{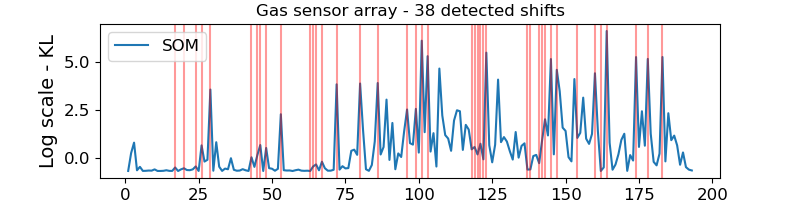}}
{\includegraphics[scale=0.7]{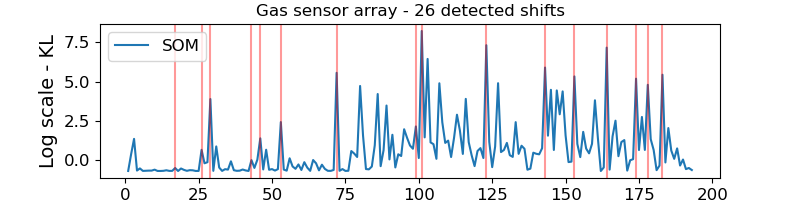}}
\caption{Gas sensor array problem. Monitoring with SOM. The figure at the top has parameters of control $\alpha=8$ and windows size equal to $10$. There are 38 detected drifts, what it is the same as the solution described in~\cite{SouzaChallenges2020}. The figure in the bottom has parameters of control $\alpha=15$ and windows size equal to $10$.
}
\label{GasSOM}
\end{figure}

\begin{figure}
    \centering
{\includegraphics[scale=0.7]{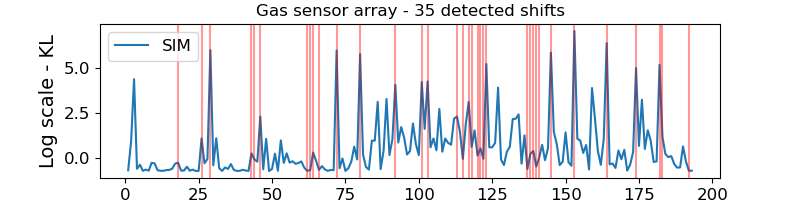}}
{\includegraphics[scale=0.7]{SIM-GASCHUNK50Alpha7Win10Detected.png}}
\caption{Gas sensor array problem. Shift monitoring with SIM. The figure in the top has control parameters $\alpha=7$ and windows size equal to $10$. There are 35 detected drifts. The figure in the bottom has parameters of control $\alpha=15$ and windows size equal to $10$. Red vertical lines show the time-stamps where our algorithm considered a significant distribution change.
}
\label{GasSIM}
\end{figure}

In addition, both baseline techniques have the additional computational cost of needing to compute the pmf estimation.
This computation is avoided in our framework because of the framework's construction; the monitoring signal follows a Gaussian distribution.
Figure~\ref{Kappa} visualizes the impact of the two parameters with respect of Kappa score in the decision rule. The left graphic depicts results obtained by SOM. The right graphic has the results obtained by SIM. 
The Kappa score is indeed a metric often used in the context of binary classification, it ranges between -1 and 1.
We can see that SOM remains stable even when parameters change, consistently achieving higher Kappa values compared to those obtained by SIM.
Figure~\ref{Incremental} presents an additional comparison between SOM and SIM. In this scenario, the results were obtained from a data stream containing chunks with samples from both the original MNIST dataset and the adversarial MNIST dataset. Consequently, this problem is more complex than monitoring streams, where the shift is precisely presented at the moment of connecting consecutive chunks.
The figure shows the different impacts of three values of $\alpha$ (specifically,~$3$, $10$, $15$) with a fixed window size of~$8$.
The framework utilizing SOM achieves high accuracy.
We visualize the results of the Gas sensor data problem in Figures~\ref{GasSOM} and~\ref{GasSIM}. 
Since this is real-world data, the shifts are not explicitly labeled. However, according to the literature, there are approximately 38 shifts~\cite{SouzaChallenges2020}.

\begin{figure}
    \centering
{\includegraphics[scale=0.4]{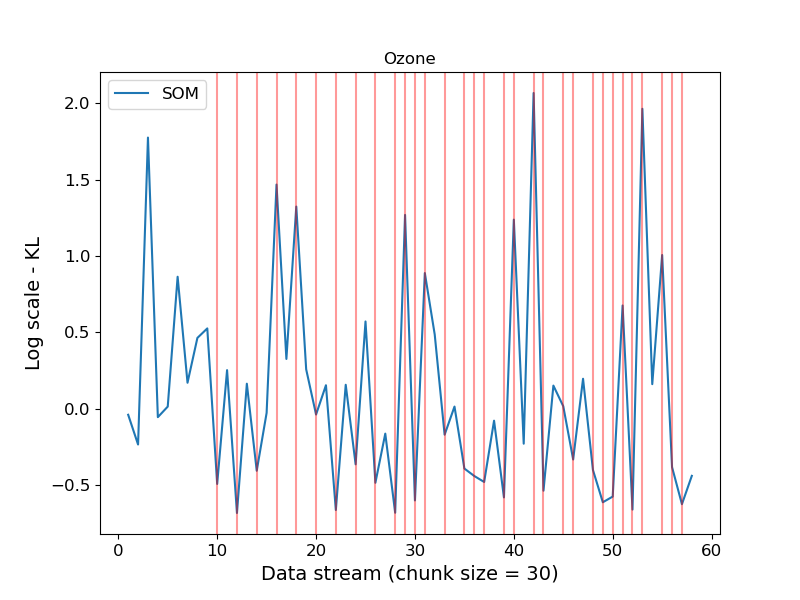}}
{\includegraphics[scale=0.4]{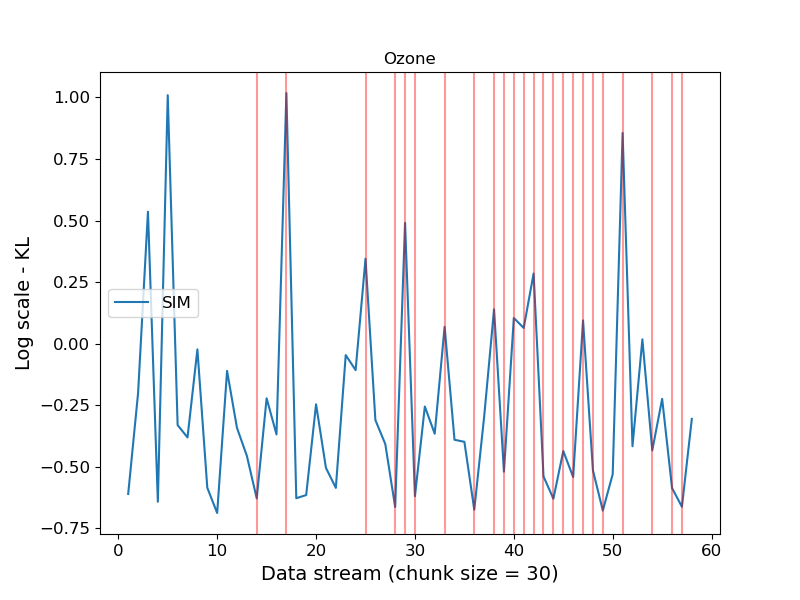}}
\caption{Ozone data set. Shift monitoring with SOM and SMM. The left figure has parameters of control $\alpha=3$ and windows size equal to $8$. 
The right figure was made with SIM, and it has parameters of control $\alpha=4$ and windows size equal to $8$. Red vertical lines show the time-stamps where our algorithm considered a significant distribution change.
}
\label{OzoneSOM}
\end{figure}

The graphics also aim to demonstrate the framework’s sensitivity to the parameters of the  $\alpha$-decision rule.
Figures~\ref{OzoneSOM} visualize the monitoring signal when the Ozone level dataset is analyzed.
This real-world problem is particularly challenging because the data is imbalanced, there are few samples, and there are many features relative to the number of samples. According to~\cite{SouzaChallenges2020}, this data stream exhibits numerous shifts (over 90 shifts). We can see that both methods are able to assess clearly discrepancies between distributions. However, it appears that the parameters chosen in the $\alpha-$decision rule strongly impact the decisions, whether they indicate a shift or not.

%
%


\section{Discussion}
\noindent{\textbf{Gaussian assumption}}. One critical aspect of our proposal is the robustness of the Gaussian assumption, needed to support the simplified computation of the KL-divergence in that case (see Subsection~\ref{Quantification}).
This relies on the Central Limit Theorem. In our case, the number of terms is given by the size of the grid of neurons, which is large enough regarding standards. The weak correlation between terms is more tricky, but the assumption often holds. This is an issue to be explored further in future studies.

\noindent{\textbf{Selected decision rule for the detection}}. Note that, the parameters of the decision rule impact in the monitoring signal too. Because when the algorithm considers that a shift occurred,  we update the weights of the topographic map. This has consequences in the monitoring signal.
However, according to the experiments, the monitoring signal is not too much sensitive to the decision rule parameters, as it is the shift detector (given by the decision rule module).
Other decision rule techniques should be explored. Threshold over peak is the easiest rule that can be integrated. The monitoring signal is univariate, therefore another possibility is to apply online outlier detection techniques.

\noindent{\textbf{Computational costs}}
Note that the proposed framework is composed of a non-linear projection (two layered NNs), a computation of a squared distance matrix ($10\times 10$ in our experiments), these distances are computed in the latent space (that it is relatively low), and the parametric estimation of KL-divergence (it requires the mean and standard deviation of a sequence with the number of samples equal to the length of the chunk). As a consequence, the method doesn't have the cost of computing histograms in large spaces that often appear in other techniques (e.g. moment tree, marginal bins~\cite{Hinder2020}). The method doesn't require either the computation of an inverse matrix as in the Kernel embedding method~\cite{Hinder2022}.
Furthermore, the distance matrix has fixed dimensions and doesn't scale with the number of samples in the chunk.

\section{Conclusions and future work}
We proposed a general system for assessing distribution shifts in high-dimensional streaming data. The approach is based on a bioinspired nonlinear projection with the property of preserving topological characteristics of the reference data, and a statistical representation of the latent space. 
The proposed method does not have any assumption about the distribution of the reference data stream. 
In addition, it can be applied to both supervised and unsupervised problems. 
We empirically study the performance of our method on three problems, and we see how the proposed system with a low budget can easily monitor the distribution changes of the input data.
Furthermore, we contrast the results of our framework  with PCA and Kernel-PCA on an annotated dataset. The results are promising and show how our proposal produce a much clearer monitoring signal than these classic techniques. 
%
We believe that the presented work offers a significant step toward the application of topology-preserving maps in the domain of distribution shifts.

In the near future, we plan to include other topology-preserving mappings (e.g. generative topographic mapping), and to explore the incorporation of other techniques for analyzing the monitoring signal.
\subsection*{Acknowledgements}
{This work was supported by the LF Experiment grant number R400-2022-1201, and it was also supported by the 22-CLIMAT-02 project entitled ``Using deep learning spatial-temporal graph models for
seasonal forecasting of extreme temperature events'' belonging to the Climate AmSud programme.}
%
%

%
\bibliographystyle{unsrt}
\bibliography{References}
\end{document}

%% file: macros.tex
\newif\ifnotes\notestrue
\def\boxnote#1#2{\ifnotes\fbox{\footnote{\ }}\ \footnotetext{ From #1:
#2}\fi}
\def\fgr#1{\boxnote{Seba}{\color{blue}#1}}
\newcommand{\msb}[1]{{\color{blue}#1}}

\newcommand{\ben}{\begin{enumerate}}
\newcommand{\een}{\end{enumerate}}
\newcommand{\predLoad}{\langle\bm{\load}\rangle}
\newcommand{\grad}{\bm{\triangledown} E}

\newcommand{\bc}{\begin{center}}
\newcommand{\ec}{\end{center}}

\newcommand{\bit}{\begin{itemize}}
\newcommand{\eit}{\end{itemize}}

\newcommand{\ds}{\displaystyle}
\newcommand{\beq}{\begin{equation}}
\newcommand{\eeq}{\end{equation}}

\newcommand{\uu}[2]{(U_{#1}^{#2})_{0,0}}
\newcommand{\spec}{\mbox{sp}}
\newcommand{\rr}{\sqrt{pq}}

\newcommand{\vak}{\va^{(k)}}
\newcommand{\vbk}{\vb^{(k)}}

\newcommand{\ppij}{p^+_{i,j}}
\newcommand{\pnij}{p^-_{i,j}}

\newcommand{\lpi}{\lambda^+_{i}}
\newcommand{\lnni}{\lambda^-_{i}}
\newcommand{\lp}{\lambda^+}
\newcommand{\lnn}{\lambda^-}
\newcommand{\load}{\varrho}
\newcommand{\loadi}{\varrho_i}

\newcommand{\Tpi}{T^+_i}
\newcommand{\Tni}{T^-_i}

\newcommand{\Tp}{T^+}
\newcommand{\Tn}{T^-}

\newcommand{\wij}{w_{i,j}}
\newcommand{\wpij}{w^+_{i,j}}
\newcommand{\wnij}{w^-_{i,j}}
\newcommand{\wpji}{w^+_{j,i}}
\newcommand{\wnji}{w^-_{j,i}}

\newcommand{\Mi}{\mathbf{M}^{\rm{in}}}
\newcommand{\Mr}{\mathbf{M}^{\rm{r}}}

\newcommand{\wn}{\mathbf{W}^{\rm{n}}}
\newcommand{\wre}{\mathbf{W}^{\rm{h}}}
\newcommand{\wi}{\mathbf{W}^{\rm{in}}}
\newcommand{\wo}{\mathbf{W}^{\rm{out}}}
\newcommand{\wfb}{\mathbf{W}^{\rm{fb}}}
\newcommand{\yt}{\y_{\text{target}}}
\newcommand{\va}{\mathbf{u}}
\newcommand{\vb}{\mathbf{b}}
\newcommand{\vx}{\mathbf{x}}
\newcommand{\vy}{\mathbf{y}}
\newcommand{\vw}{\mathbf{W}}
\newcommand{\vat}{\mathbf{a}(t)}
\newcommand{\vbt}{\mathbf{b}(t)}
\newcommand{\vs}{\mathbf{s}}
\newcommand{\Na}{n}
\newcommand{\Ny}{o}
\newcommand{\Nx}{p}
\newcommand{\R}{\mathbb{R}}
\newcommand{\vg}{\boldsymbol\gamma}
\newcommand{\cdI}{{I}}
\newcommand{\cdO}{{O}}
\newcommand{\X}{\mathbf{X}}
\newcommand{\Y}{\mathbf{Y}}
\newcommand{\Prob}{\mathds{P}}
\newcommand{\N}{\mathds{N}}
\newcommand{\gradientLb}{\bigtriangledown L(\bm{\beta})}

\newcommand{\SB}[1]{\textbf{\sf #1}}
\newcommand{\fpNoParen}[2]{\displaystyle{\frac{\partial{#1}}{\partial{#2}}}}
\newcommand{\fp}[2]{\displaystyle{\frac{\partial}{\partial{#2}}\bigg({#1}\bigg)}}

\newcommand{\indice}[1]{#1\index{\sc{#1}}}
\newcommand{\indicat}[1]{\upharpoonleft\!\mid_{(#1)}}